\def\year{2021}\relax
\documentclass[letterpaper]{article} 
\usepackage{aaai21}  
\usepackage{times}  
\usepackage{helvet} 
\usepackage{courier}  
\usepackage[hyphens]{url}  
\usepackage{graphicx} 
\usepackage{natbib}  
\usepackage{caption} 
\usepackage{grffile}
\usepackage[export]{adjustbox}
\usepackage{amsfonts}       
\usepackage{amsmath}        
\usepackage{amssymb}        
\usepackage{array}          
\usepackage{bm}             
\usepackage{booktabs}       
\usepackage{forest}         
\usepackage{graphicx}       
\usepackage{hyperref}       
\usepackage{ifthen}         
\usepackage{makecell}       
\usepackage{mathrsfs}       
\usepackage{microtype}      
\usepackage{multirow}       
\usepackage{pgfplots}       
\pgfplotsset{compat=1.14}
\usepackage{setspace}       %
\usepackage{subcaption}     
\usepackage{subdepth}       
\usepackage{tikz}           
\usepackage{verbatim}       
\usepackage{url}            
\usepackage{xfrac}          

\long\def\ignorethis#1{}

\newcommand{\eg}{\textit{e.g.,}~}
\newcommand{\ie}{\textit{i.e.,}~}

\definecolor{darkgreen}{RGB}{0,125,0}
\definecolor{orange}{RGB}{255,124,84}
\definecolor{turquoise}{rgb}{0.19, 0.84, 0.78}
\newcounter{NoteCounter}

\definecolor{shadecolor}{gray}{0.9}

\usepackage{graphicx}
\usepackage[font=small,labelfont=bf]{caption}
\usepackage[format=hang]{subcaption}

\usepackage[algo2e,algoruled,vlined,noend,ruled]{algorithm2e}  
\AtBeginEnvironment{algorithm2e}{\setstretch{1.05}}            
\SetAlFnt{\footnotesize}
\SetAlCapFnt{\footnotesize}
\SetAlCapNameFnt{\footnotesize}

\usepackage{listings}
\usepackage{fancyvrb}
\fvset{fontsize=\footnotesize}
\usepackage{etoolbox}

\usepackage[all]{hypcap}

\newcommand{\figref}[1]{\hyperref[fig:#1]{Figure~\ref{fig:#1}}}
\newcommand{\twofigref}[2]{Figures~\hyperref[fig:#1]{\ref{fig:#1}} and~\hyperref[fig:#2]{\ref{fig:#2}}}

\newcommand{\tabref}[1]{\hyperref[tab:#1]{Table~\ref{tab:#1}}}
\newcommand{\twotabref}[2]{Tables~\hyperref[tab:#1]{\ref{tab:#1}} and~\hyperref[tab:#2]{\ref{tab:#2}}}

\renewcommand{\eqref}[1]{\hyperref[eq:#1]{Equation~(\ref{eq:#1})}}
\newcommand{\shorteqref}[1]{\hyperref[eq:#1]{(\ref{eq:#1})}}
\newcommand{\twoeqref}[2]{Equations~\hyperref[eq:#1]{\ref{eq:#1}} and~\hyperref[eq:#2]{\ref{eq:#2}}}

\newcommand{\secref}[1]{\hyperref[sec:#1]{Section~\ref{sec:#1}}}
\newcommand{\twosecref}[2]{Sections~\hyperref[sec:#1]{\ref{sec:#1}} and~\hyperref[sec:#2]{\ref{sec:#2}}}

\newcommand{\appref}[1]{\hyperref[app:#1]{Appendix~\ref{app:#1}}}
\newcommand{\thmref}[1]{\hyperref[thm:#1]{Theorem~(\ref{thm:#1})}}
\newcommand{\algref}[1]{\hyperref[alg:#1]{Algorithm~\ref{alg:#1}}}
\renewcommand{\subref}[1]{\hyperref[alg:#1]{Subroutine~\ref{alg:#1}}}

\makeatletter
\let\UrlSpecialsOld\UrlSpecials
\def\UrlSpecials{\UrlSpecialsOld\do\/{\Url@slash}\do\_{\Url@underscore}}%
\def\Url@slash{\@ifnextchar/{\kern-.11em\mathchar47\kern-.2em}%
    {\kern-.0em\mathchar47\kern-.08em\penalty\UrlBigBreakPenalty}}
\def\Url@underscore{\nfss@text{\leavevmode \kern.06em\vbox{\hrule\@width.3em}}}
\makeatother

\usepackage[
  acronym,
  smallcaps,
  nowarn,
  section,
  nonumberlist
]{glossaries-extra}
\glsdisablehyper{}

\lstdefinestyle{mystyle}{
    commentstyle=\color{OliveGreen},
    numberstyle=\tiny\color{black!60},
    stringstyle=\color{BrickRed},
    basicstyle=\ttfamily\scriptsize,
    breakatwhitespace=false,
    breaklines=true,
    captionpos=b,
    keepspaces=true,
    numbers=none,
    numbersep=5pt,
    showspaces=false,
    showstringspaces=false,
    showtabs=false,
    tabsize=2
}
\tikzset{every picture/.style=remember picture}

\setabbreviationstyle[long]{long}             
\setabbreviationstyle[long-noshort]{long-em}  

\newabbreviation[category=long-noshort]{babyai}{}{BabyAI}
\newabbreviation[category=long-noshort]{ch}{}{cognitive hierarchy}
\newabbreviation[category=long-noshort]{da}{}{domain adaptation}
\newabbreviation[category=long-noshort]{eb}{}{empirical Bayes}
\newabbreviation[category=long-noshort]{es}{}{early stopping}
\newabbreviation[category=long-noshort]{fa}{}{fast adaptation}
\newabbreviation[category=long-noshort]{fisher}{}{Fisher information matrix}
\newabbreviation[category=long-noshort]{hbm}{}{hierarchical Bayesian model}
\newabbreviation[category=long-noshort]{ho}{}{hyperparameter optimization}
\newabbreviation[category=long-noshort]{iid}{}{independent and identically distributed}
\newabbreviation[category=long-noshort]{ib}{}{inductive bias}
\newabbreviation[category=long-noshort]{gb}{}{gradient-based}
\newabbreviation[category=long-noshort]{gen}{}{generative}
\newabbreviation[category=long-noshort]{joint}{}{joint training}
\newabbreviation[category=long-noshort]{jt}{}{joint training}
\newabbreviation[category=long-noshort]{kf}{}{Kronecker factor}
\newabbreviation[category=long-noshort]{laplace}{}{the Laplace approximation}
\newabbreviation[category=long-noshort]{ltl}{}{learning-to-learn}
\newabbreviation[category=long-noshort]{mse}{}{mean-squared error}
\newabbreviation[category=long-noshort]{mini}{}{\textit{mini}ImageNet}
\newabbreviation[category=long-noshort]{ml}{}{meta-learning}
\newabbreviation[category=long-noshort]{mnist}{}{MNIST}
\newabbreviation[category=long-noshort]{np}{}{non-parametric}
\newabbreviation[category=long-noshort]{omni}{}{Omniglot}
\newabbreviation[category=long-noshort]{p}{}{parametric}
\newabbreviation[category=long-noshort]{permmnist}{}{permuted MNIST}
\newabbreviation[category=long-noshort]{qfunc}{}{$Q$-function}
\newabbreviation[category=long-noshort]{rnn}{}{recurrent neural network}
\newabbreviation[category=long-noshort]{splitcifar}{}{split CIFAR-10/100}
\newabbreviation[category=long-noshort]{tl}{}{transfer learning}
\newabbreviation[category=long-noshort]{tom}{}{theory of mind}
\newabbreviation[category=long-noshort]{side-info}{}{\emph{contextual adaptation}}

\newabbreviation[category=long-short]{a3c}{A3C}{asynchronous advantage actor-critic}
\newabbreviation[category=long-short]{ban}{BAN}{born again network}
\newabbreviation[category=long-short]{bn}{BN}{batch normalization}
\newabbreviation[category=long-short]{bnp}{BNP}{Bayesian non-parametric}
\newabbreviation[category=long-short]{celeba}{CelebA}{CelebFaces attributes}
\newabbreviation[category=long-short]{coma}{COMA}{counterfactual multi-agent}
\newabbreviation[category=long-short]{cnn}{CNN}{convolutional neural network}
\newabbreviation[category=long-short]{crp}{CRP}{Chinese restaurant process}
\newabbreviation[category=long-short]{cub}{CUB-200}{Caltech-UCSD Birds 200}
\newabbreviation[category=long-short]{fscub}{few-shot CUB-200}{few-shot Caltech-UCSD Birds 200}
\newabbreviation[category=long-short]{dpmm}{DPMM}{Dirichlet process mixture model}
\newabbreviation[category=long-short]{dqn}{DQN}{deep Q-network}
\newabbreviation[category=long-short]{drl}{DRL}{deep reinforcement learning}
\newabbreviation[category=long-short]{em}{EM}{expectation maximization}
\newabbreviation[category=long-short]{film}{FiLM}{feature-wise linear modulation}
\newabbreviation[category=long-short]{gae}{GAE}{generalized advantage estimation}
\newabbreviation[category=long-noshort]{gbml}{GBML}{gradient-based meta-learning}
\newabbreviation[category=long-short]{hb}{HB}{hierarchical Bayes}
\newabbreviation[category=long-short]{icarl}{iCaRL}{incremental classifier and representation learning}
\newabbreviation[category=long-short]{icm}{ICM}{iterated conditional modes}
\newabbreviation[category=long-short]{ilsvrc}{ILSVRC}{ImageNet Large Scale Visual Recognition Challenge}
\newabbreviation[category=long-short]{ipomdp}{I-POMDP}{interactive Markov decision process}
\newabbreviation[category=long-short]{it}{IT}{iterated learning}
\newabbreviation[category=long-short]{kfac}{K-FAC}{Kronecker-factored approximate curvature}
\newabbreviation[category=long-short]{leo}{LEO}{latent embedding optimization}
\newabbreviation[category=long-short]{llama}{LLAMA}{lightweight Laplace approximation for meta-adaptation}
\newabbreviation[category=long-short]{lola}{LOLA}{learning with opponent-learning awareness}
\newabbreviation[category=long-short]{meta-rl}{meta-RL}{meta-reinforcement learning}
\newabbreviation[category=long-short]{mle}{MLE}{maximum likelihood}
\newabbreviation[category=long-short]{map}{MAP}{\textit{maximum a posteriori}}
\newabbreviation[category=long-short]{maml}{MAML}{model-agnostic meta-learning}
\newabbreviation[category=long-short]{mppi}{MPPI}{model-predictive path integral}
\newabbreviation[category=long-short]{marl}{MARL}{multi-agent reinforcement learning}
\newabbreviation[category=long-short]{mdp}{MDP}{Markov decision process}
\newabbreviation[category=long-short]{mcmc}{MCMC}{Markov chain Monte Carlo}
\newabbreviation[category=long-short]{mn}{MN}{matching network}
\newabbreviation[category=long-short]{mpc}{MPC}{model predictive control}
\newabbreviation[category=long-short]{mujoco}{MuJoCo}{multi-joint dynamics with contact}
\newabbreviation[category=long-noshort]{metaworld}{}{Meta-World}
\newabbreviation[category=long-noshort]{ml1}{}{ML1}
\newabbreviation[category=long-short]{nll}{NLL}{negative log likelihood}
\newabbreviation[category=long-short]{pomdp}{POMDP}{partially observable Markov decision process}

\newabbreviation[category=long-short]{rl}{RL}{reinforcement learning}
\newabbreviation[category=long-short]{sl}{SL}{supervised learning}
\newabbreviation[category=long-short]{rsa}{RSA}{rational speech acts}
\newabbreviation[category=long-short]{sgd}{SGD}{stochastic gradient descent}
\newabbreviation[category=long-short]{si}{SI}{synaptic intelligence}
\newabbreviation[category=long-short]{tadam}{TADAM}{task-dependent adaptive metric}
\newabbreviation[category=long-short]{td}{TD}{temporal difference}
\newabbreviation[category=long-short]{vcl}{VCL}{variational continual learning}

\newabbreviation[category=long-noshort]{half-cheetah}{}{\texttt{HalfCheetahRandDirec}}
\newabbreviation[category=long-noshort]{walker-2d}{}{\texttt{Walker2DRandDirec}}
\newabbreviation[category=long-noshort]{ant-goal}{}{\texttt{AntRandGoal}}
\newabbreviation[category=long-noshort]{reach}{}{\texttt{reach-v1}}
\newabbreviation[category=long-noshort]{push-wall}{}{\texttt{push-wall-v1}}
\newabbreviation[category=long-noshort]{push}{}{\texttt{push-v1}}
\newabbreviation[category=long-noshort]{pick-place-wall}{}{\texttt{pick-place-wall-v1}}
\newabbreviation[category=long-noshort]{door-lock}{}{\texttt{door-lock-v1}}
\newabbreviation[category=long-noshort]{door-unlock}{}{\texttt{door-unlock-v1}}
\newabbreviation[category=long-noshort]{soccer}{}{\texttt{soccer-v1}}
\newabbreviation[category=long-noshort]{basket-ball}{}{\texttt{basket-ball-v1}}

\newabbreviation[category=long-noshort]{ppo}{\textsc{PPO}}{proximal policy optimization}
\newabbreviation[category=long-short]{promp}{\textsc{ProMP}}{proximal meta-policy search}
\newabbreviation[category=long-noshort]{promp-concat}{}{\textsc{ProMP-concat}}
\newabbreviation[category=long-noshort]{promp-static}{}{\textsc{ProMP-static}}
\newabbreviation[category=long-noshort]{maml-concat}{}{\textsc{MAML-concat}}
\newabbreviation[category=long-noshort]{maml-static}{}{\textsc{MAML-static}}
\newabbreviation[category=long-short]{maml-context}{\textsc{MLTI}}{\emph{meta-learning with task information}}
\newabbreviation[category=long-short]{promp-context}{\textsc{MLTI}}{\emph{meta-learning with task information}}

\newcommand{\bd}[1]{\partial{\kern0pt#1}}
\newcommand{\interior}[1]{{\kern0pt#1\kern1pt}^{\mathrm{o}}}

\newcommand{\tr}{^\mathrm{T}}

\newcommand{\sample}{\sim}
\newcommand{\given}{\;\vert\;}

\newcommand{\boldvar}[1]{\bm{#1}}

\newcommand{\boldSigma}{\boldvar{\Sigma}}

\newcommand{\boldtheta}{\boldvar{\theta}}

\newcommand{\boldphi}{\boldvar{\phi}}

\newcommand{\boldmu}{\boldvar{\mu}}

\DeclareMathAlphabet{\mathsfit}{\encodingdefault}{\sfdefault}{m}{sl}
\SetMathAlphabet{\mathsfit}{bold}{\encodingdefault}{\sfdefault}{bx}{n}

\newcommand{\mean}{\boldmu}
\newcommand{\covariance}{\boldSigma}

\newcommand{\MVN}[3]{{\frac {\exp \left(-{\frac {1}{2}}({\mathbf {x} }-{\mean})\tr }{\covariance}^{-1}(\bx-{\mean})\right)}{\sqrt {|2\pi {\covariance|}}}}

\newcommand{\policy}{\mathbf{\pi}}

\newcommand{\mdpdistribution}{\rho}

\usepackage{pgfplots}
\usepackage{pgfplotstable}
\pgfplotsset{compat=1.3}
\usepackage{tikz}

\usetikzlibrary{matrix}
\usetikzlibrary{bayesnet}
\usetikzlibrary{calc}
\usetikzlibrary{decorations.pathmorphing}
\usetikzlibrary{positioning}
\usetikzlibrary{pgfplots.groupplots}
\usetikzlibrary{spy}
\usepgfplotslibrary{fillbetween}

\pgfdeclarelayer{edgelayer}
\pgfdeclarelayer{nodelayer}
\pgfsetlayers{edgelayer,nodelayer,main}

\definecolor{grey}{rgb}{0.749,0.749,0.749}

\tikzset{>=latex}

\tikzstyle{tensor} = [circle,fill=white,draw=black]
\tikzstyle{op}     = [rectangle,fill=grey,draw=black]

\newcommand{\metaparams}{\boldtheta}
\newcommand{\taskparams}{\boldphi}

\newcommand{\task}{\mathcal{T}}

\newcommand{\taskidx}[1]{_{#1}}

\newcommand{\val}{\scriptstyle\text{val}}

\newcommand{\context}{\mathbf{c}}
\newcommand{\contextparams}{\psi}

\newcommand{\errorband}[6]{
    \addplot [name path=pluserror,draw=none,no markers,forget plot]
      table [col sep=comma, x={#2},y expr=\thisrow{#3}+\thisrow{#4}] {#1};
  
    \addplot [name path=minuserror,draw=none,no markers,forget plot]
       table [col sep=comma, x={#2},y expr=\thisrow{#3}-\thisrow{#4}] {#1};
  
    \addplot [forget plot,fill=#5,opacity=#6]
      fill between[on layer={},of=pluserror and minuserror];
  
    \addplot+[#5,thick,no markers]
      table [col sep=comma, x={#2} ,y={#3}] {#1};
}

\usepackage{amsmath,amsfonts,bm}

\def\figref#1{figure~\ref{#1}}

\def\twofigref#1#2{figures \ref{#1} and \ref{#2}}

\def\secref#1{section~\ref{#1}}

\def\eqref#1{equation~\ref{#1}}

\def\algref#1{algorithm~\ref{#1}}

\def\1{\bm{1}}

\DeclareMathAlphabet{\mathsfit}{\encodingdefault}{\sfdefault}{m}{sl}
\SetMathAlphabet{\mathsfit}{bold}{\encodingdefault}{\sfdefault}{bx}{n}

\usepackage[utf8]{inputenc} 
\usepackage[T1]{fontenc}    
\usepackage{hyperref}       
\usepackage{url}            
\usepackage{booktabs}       
\usepackage{amsfonts}       
\usepackage{nicefrac}       
\usepackage{microtype}      
\usepackage{graphicx}
\usepackage{hyperref}
\usepackage{pifont}         
\usepackage{arydshln}       
\usepackage{xcolor}         
\usepackage{enumitem}         
\usepackage[switch]{lineno}

\usepackage{filecontents,tikz}
\usepackage{tikzscale}
\usetikzlibrary{external}

\newcommand{\cmark}{\ding{51}}%
\newcommand{\xmark}{\ding{55}}%

\usepackage{wrapfig}

\usepackage{caption}
\usepackage{subcaption}
\title{Connecting context-specific adaptation in humans to meta-learning}

\author{
  Rachit Dubey*\textsuperscript{\rm 1},
  Erin Grant*\textsuperscript{\rm 2},
  Michael Luo*\textsuperscript{\rm 2},
  Karthik Narasimhan\textsuperscript{\rm 1}, 
  Thomas L.~Griffiths\textsuperscript{\rm 1} 
}

\affiliations{
  Princeton University\textsuperscript{\rm 1},
  UC Berkeley\textsuperscript{\rm 2} \\
  \small{
  \texttt{\{rdubey,tomg\}@princeton.edu},
  \texttt{\{eringrant,michael.luo\}@berkeley.edu},
  \texttt{karthikn@cs.princeton.edu}
  }
}

\pgfplotsset{
  compat=newest,
  /pgfplots/legend image code/.code={%
    \draw[mark repeat=1,mark phase=1,#1]
      plot coordinates {
        (0cm,0cm)
        (0.2cm,0cm)
      };
  },
}

\pgfplotsset{
  avg-loss/.style={
    legend style={
      draw=none,
      font=\tiny,
      align=left,
      column sep=3pt,
      nodes={right},
      legend style={at={(1.2,.5)},anchor=west},
    },
    width=50pt,
    height=40pt,
    y label style={align=center,font=\bfseries\scriptsize},
    x label style={font=\bfseries\scriptsize},
    tick label style={font=\scriptsize},
    enlarge x limits=false,
    enlarge y limits=false,
    scale only axis,
    scaled x ticks=true,
  },
}

\pgfplotsset{
  acc-over-shot/.style={
    legend pos=outer north east,
    legend style={
      font=\tiny,
      align=left,
      column sep=3pt,
      nodes={right},
    },
    y label style={yshift=-2pt,align=center,font=\bfseries\scriptsize},
    x label style={font=\bfseries\scriptsize},
    tick label style={font=\tiny},
    enlarge x limits=false,
    scale only axis,
    width=90pt,
    height=67.5pt,
    xlabel={support set size (\emph{shot})},
    ylabel={average accuracy},
    ytick={0,50,100},
    ymin=0, ymax=100,
    xtick={1,2,3,4,5},
    xmin=1, xmax=5,
  },
}

\pgfplotsset{
  avg-return/.style={
    legend style={
      font=\tiny,
      align=left,
      column sep=3pt,
      nodes={right},
      legend style={at={(1.1,.5)},anchor=west},
      draw=none,
    },
    y label style={align=center,font=\bfseries\scriptsize,yshift=-5pt},
    x label style={font=\bfseries\scriptsize},
    tick label style={font=\tiny},
    enlarge x limits=false,
    grid=major,
    scale only axis,
    width=50pt,
    height=40pt,
    scaled x ticks=false,
    tick scale labels in axis labels/.code={
            \pgfkeysgetvalue{/pgfplots/xtick scale label code/.@cmd}\temp
            \pgfkeyslet{/pgfplots/xtick scale label code orig/.@cmd}\temp
            \pgfkeysalso{
                xtick scale label code/.code={
                    \gdef\xTickScale{##1}
                },
                every axis/.append style={
                    tick scale binop={},
                    xlabel/.add = {}{
                        (\pgfplotsset{xtick scale label code orig=\xTickScale}\,m)
                    },
                },
            }
        },
  },
}

\begin{document}

\newcommand{\erin}[1]{({\color{orange}{Erin: #1}})}
\newcommand{\rachit}[1]{({\color{blue}{Rachit: #1}})}
\newcommand{\michael}[1]{({\color{cyan}{Michael: #1}})}
\newcommand{\karthik}[1]{({\color{magenta}{Karthik: #1}})}
\newcommand{\tom}[1]{({\color{red}{Tom: #1})}}
\newcommand{\todo}[1]{({\color{red}{#1)}}}

\maketitle

\begin{abstract}
  Cognitive control, the ability of a system to adapt to the demands of a task, is an integral part of cognition.
A widely accepted fact about cognitive control is that it is context-sensitive: Adults and children alike infer information about a task's demands from contextual cues and use these inferences to learn from ambiguous cues.
However, the precise way in which people use contextual cues to guide adaptation to a new task remains poorly understood. This work connects the context-sensitive nature of cognitive control to a method for meta-learning with context-conditioned adaptation.
We begin by identifying an essential difference between human learning and 
 current approaches to
meta-learning: In contrast to humans, existing meta-learning algorithms do not make use of task-specific contextual cues but instead rely exclusively on online feedback in the form of task-specific labels or rewards.
To remedy this, we introduce a framework for using contextual information about a task to guide the initialization of task-specific models before adaptation to online feedback.
We show how context-conditioned meta-learning can capture human behavior in a cognitive task and how it can be scaled to improve the speed of learning in various settings, including few-shot classification and low-sample reinforcement learning. 
Our work demonstrates that guiding meta-learning with task information can capture complex, human-like behavior, thereby deepening our understanding of cognitive control.

\end{abstract}


\section{Introduction}
\label{sec:intro}

Flexibility is one of the defining features of cognition. Any intelligent organism must be able to adapt its behavior to continually changing and evolving environmental and task demands~\citep{braver2009flexible}. The processes behind such adaptability are collectively referred to as \emph{cognitive control}~\citep{cohen2000anterior, botvinick2001conflict, botvinick2014computational}, and a primary goal of modern cognitive psychology and neuroscience involves understanding the mechanisms that underlie cognitive control in humans~\citep{barcelo2006task}.

A notable feature of cognitive control is the ability to derive complex rules from contextual cues~\citep{monsell1996control, salinas2004fast, dosenbach2006core, sakai2008task, collins2013cognitive}. As an example, consider a child raised in a bilingual environment with each parent speaking a different language. Upon learning that each parent speaks a different language, the child may come to expect that depending on the speaker (the context), the same object (the stimulus) will be labeled using different words (the response)~\citep{werchan20158}. In this manner, contextual information such as visual or linguistic cues enables adults and children alike to recognize the underlying structure of a new problem they face, which, in turn, enables them to decide on a strategy for interaction within the novel context ~\citep{collins2012reasoning, collins2013cognitive, werchan20158}.

Although it is well established that context-dependent adaptation is vital for flexible behavior, the computational mechanisms underlying how humans use contextual information to guide learning in a new situation are still poorly understood. While recent computational works have shed essential insights into understanding these mechanisms in simplified settings~\citep{collins2013cognitive, eckstein2020computational}, we lack computational models that can be scaled up to more realistic tasks.

In the present work, we offer a new perspective by proposing that context-dependent adaptation can be explained within a context-conditioned meta-learning framework. In standard meta-learning, a meta-learned global model determines the initialization of task-specific models, which are subsequently adapted to online feedback from each task.  Here, we propose \gls{maml-context}, in which contextual cues about task structure--termed \emph{task information}-- guide the initialization of task-specific models, enabling the meta-learned prior over task structures to be informed by task information, similar to how human learning is guided by context. 

We implement \gls{maml-context} by augmenting a gradient-based meta-learning algorithm~\citep{finn2017model} with a \emph{context network} that learns the relationship between task information and the initialization of task-specific models. We first use this implementation to demonstrate that the \gls{maml-context} framework can capture the context-sensitivity of human behavior in a simple but well-studied cognitive control task. We then shift our focus to larger-scale simulations, where we demonstrate competitive performance against several baselines on supervised and reinforcement learning tasks. Our work thus contributes a framework for understanding key aspects of human adaptability and a cognitively-inspired algorithm that is competitive in realistic settings.

\section{Background}
\label{ref:background}

\subsubsection{Computational accounts of context-specific adaptation in humans.} 
Although the importance of contextual cues in guiding human flexibility is well-established, very little work has looked into how contextual information guides such adaptability. Recent computational works have made progress towards understanding these mechanisms by suggesting that context-specific adaptation can be modeled using nonparametric Bayesian methods~\citep{collins2013cognitive} as well as hierarchical reinforcement learning~\citep{eckstein2020computational}. However, one limitation of these works is that the tasks modeled using these frameworks are relatively simple compared to the problems faced by humans. This limitation restricts our understanding of context-sensitive adaptation as we do not have models that can capture our everyday flexibility and adaptability. Despite this limitation, a critical insight from these models is that they suggest that the learning processes involved in cognitive control occur at multiple levels of abstraction in that prior knowledge and cognitive control constrain the lower-level, stimulus-response learning~\citep{collins2018learning}. We take this insight as the motivation to pursue modeling context-specific adaptation under a \emph{meta-learning} framework, which realizes an analogous hierarchical decomposition of learning.

\subsubsection{Meta-learning.}

Meta-learning aims to learn a model suitable for a distribution of tasks, which subsequently enables few-shot adaptation to new tasks sampled from the same distribution~\citep{schmidhuber1987evolutionary,bengio1992optimization,thrun1998lifelong}, formulated in recent
works as the learning of global parameters that are shared
between independent, task-specific models \citep{finn2017model, vinyals2016matching}. While meta-learning algorithms can capture some elements of human adaptability (such as the ability to learn from very few examples), standard formulations of meta-learning are not sufficient to capture context-sensitive adaptation. This is because popular meta-learning approaches~\cite[\eg][]{vinyals2016matching, finn2017model,snell2017prototypical} and their derivatives learn in the absence of abstract task information by treating each task as a uniformly random draw from an underlying task distribution and do not use context to prime their learning. 

\subsubsection{Context-conditioning in meta-learning.}

Recent works have explored augmenting meta-learning with conditioning information by modifying the meta-learner architectures to encode task-specific data into a latent task representation \citep{oreshkin2018tadam, pahde2018cross, vuorio2018toward, chen2019adaptive, lee2018gradient, lee2019learning, baik2019learning, lan2019meta, yoon2019tapnet}. Analogous to the way learning loops occurring between abstract contexts and high-level rules constrain the lower-level learning loop in the brain, in these frameworks, outer learning loop between latent task representation and high-level rules constrain the inner learning loop. 

However, one important distinction between context-conditioning meta-learning and context-specific human adaptation is that the former produces the task encoding using the support set \ie  using the \emph{same} data over which the meta-learning objective is defined. For instance, \citep{oreshkin2018tadam, vuorio2018toward, baik2019learning, lan2019meta, lee2018gradient} use a conditioning network to infer information about the task, but they do so without employing external contextual information. Similarly, \cite{lee2018gradient} propose a meta-learning model that uses a transformation network to augment the base network with an implicit conditional network as a linear transformation on the weights but uses the same data as the base network. \cite{pahde2018cross, chen2019adaptive} also use contextual information at the instance or class level without any conditioning network. \cite{yoon2019tapnet} linearly transform feature embeddings with a task-specific projection but does not employ contextual information or a conditioning network. This means that while context-conditioning meta-learning enables efficient few-shot learning, it cannot fully capture and explain context-sensitive adaptation in humans. 

\section{The present research}

In this work, we consider meta-learning as a useful starting point towards modeling context-sensitive adaptation in humans. However, as noted previously, unlike humans, standard formulations of meta-learning do not employ contextual cues, and only in some cases, infer a task representation from task-specific data.

To account for human behavior, we instead propose to use contextual cues to \emph{guide} meta-learning. Unlike prior works on meta-learning, we produce a task representation from the \emph{extra} available contextual information and focus on the utility of this information in structuring learning at a higher level of abstraction rather than the increased expressiveness that architectural modifications bring. This structure is motivated by human learning, in which contextual cues serve to inform a prior about the task structure at hand, which then enables rapid adaptation to novel contexts. Our experiments show that this task-specific contextual-adaptation can not only capture human behavior but also improve the speed of learning of meta-learning in supervised and reinforcement learning tasks.

Our key contributions are as follows. First, to explain context-sensitive adaptation in humans, we introduce a framework that uses task information to guide meta-learning. Second, we demonstrate that our framework can successfully capture human behavior in a well-known cognitive control task. Modeling human behavior in this task allows us to understand important aspects of human flexibility and cognitive control. Third, and unusually for a cognitive modeling framework, we show that models implemented in our framework can outperform competitive baselines in more complex problem domains such as \gls{celeba} and \gls{metaworld}. Thus, our work also contributes towards developing a cognitively inspired meta-learning framework that can be applied to more realistic problem domains.

\begin{figure}[t]
  \begin{center}
    \includegraphics[width=0.3\textwidth]{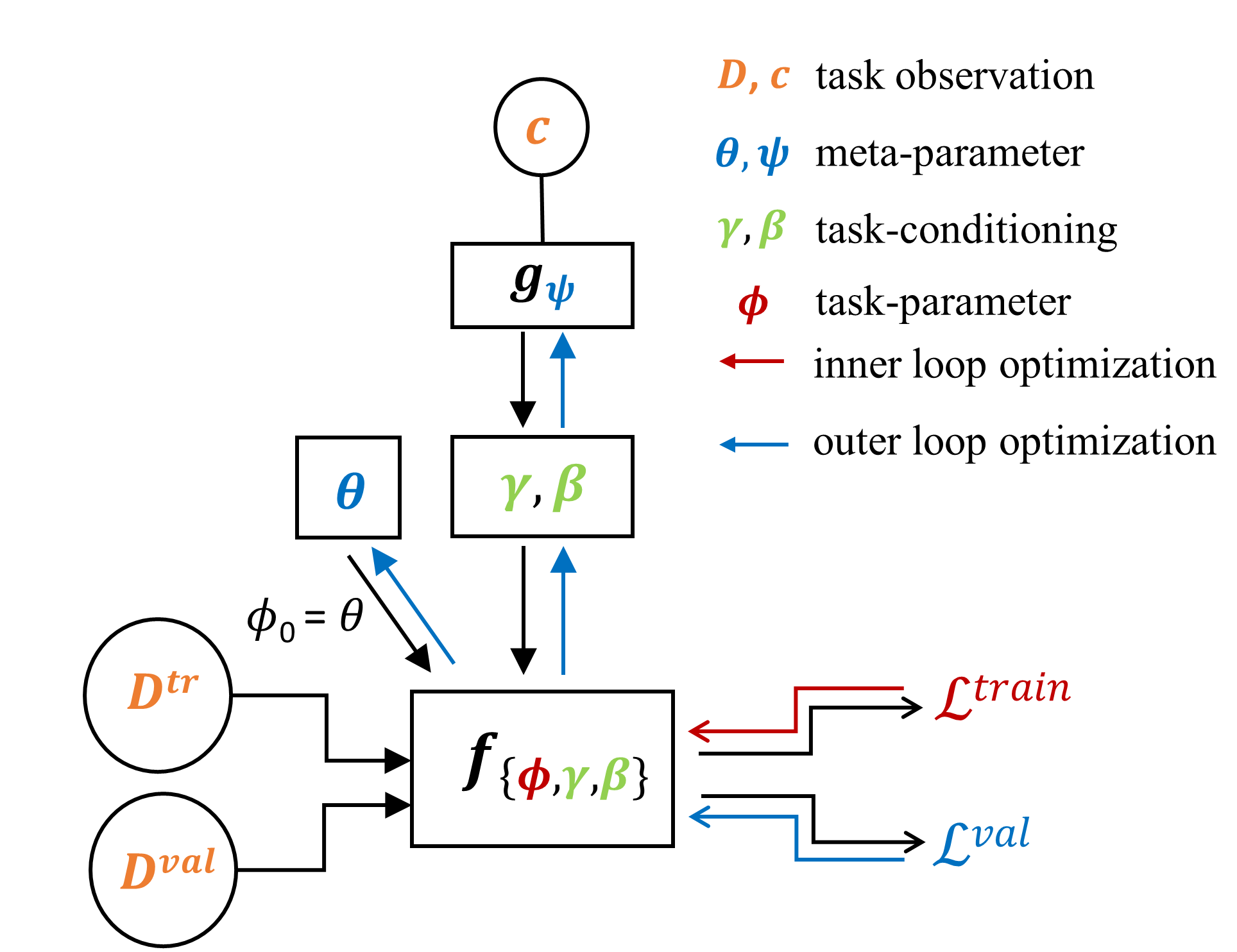}
  \end{center}
  \caption{We use task-specific contextual information by learning an initialization of the task-specific model parameters $\taskparams$ as a function of the task information $\context$ to leverage meta-learning. This framework of guiding meta-learning with task information allows us to explain context-specific adaptation in humans.}
\label{fig:architecture}
\end{figure}

\begin{figure*}[t]
\begin{subfigure}[b]{.2\linewidth}

  \includegraphics[height=70pt]{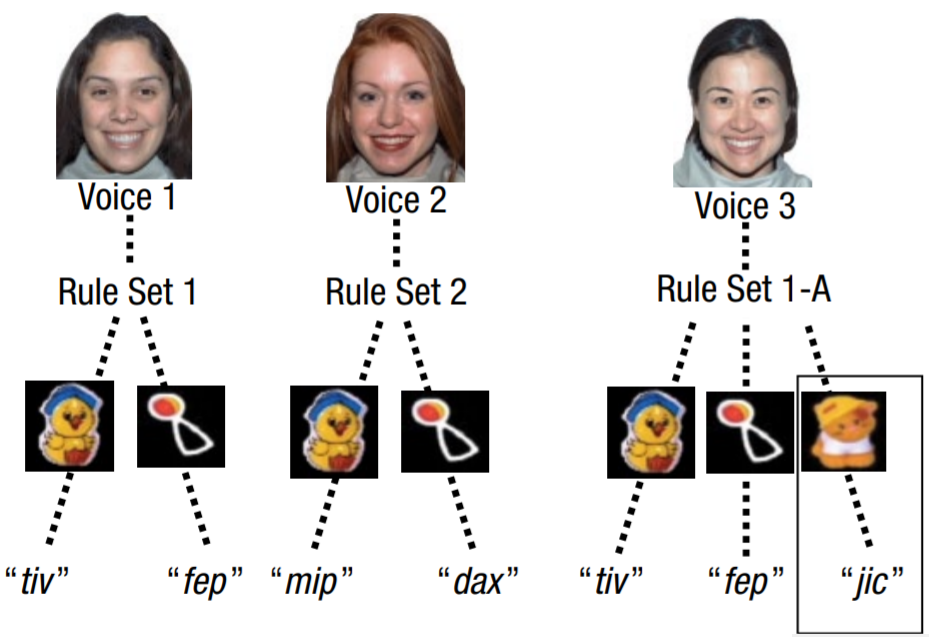}

\caption{
  The learning task from \cite{werchan20158,werchan2016role}: Three speakers use two rule sets (object-word mapping) to label three stimuli, the third object is labeled only once by a speaker whose labeling are consistent with one of the two rule sets.
}
\label{fig:cognition1}
\end{subfigure}
\hfill
\begin{subfigure}[b]{.22\linewidth}
\begin{tikzpicture}%
  \begin{axis}[%
    avg-loss,
    ylabel={training loss},
    xlabel={\scriptsize{meta-training iterations}},
    xmin=0,xmax=98,
    y tick label style={/pgf/number format/1000 sep=\,},
    legend style={
      legend style={at={(1.0,.5)},anchor=west},
    }
  ]

  \errorband{data/cognitive/maml.csv}{iteration}{mean}{std}{black}{0.4}
  \addlegendentry{\acrshort{maml} \\ (base)}

  \errorband{data/cognitive/mlti.csv}{iteration}{mean}{std}{orange}{0.4}
  \addlegendentry{\acrshort{maml-context} \\ \textbf{(ours)}}

  \end{axis}
\end{tikzpicture}%

\caption{
  Learning curves ($n$ = 5 random seeds) for \acrshort{maml} and \acrshort{maml-context} on a reproduction of the learning task from \cite{werchan20158,werchan2016role}.
  \acrshort{maml-context} makes use of task information 
  to learn more rapidly and with much lower variance than \acrshort{maml}.
}
\label{fig:cognition2}
\end{subfigure}
\hfill
\begin{subfigure}[b]{.16\linewidth}

\includegraphics[height=55pt]{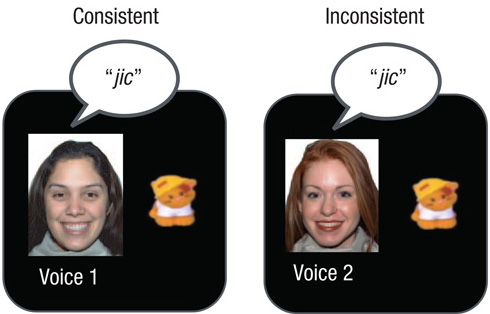}

\caption{
  The inference test:  In the (in)consistent condition, the label is given by the speaker with a rule set (in)consistent with that of the speaker who named the third stimulus in (a).
}
\label{fig:cognition3}
\end{subfigure}
\hfill
\begin{subfigure}[b]{.15\linewidth}

\includegraphics[height=55pt]{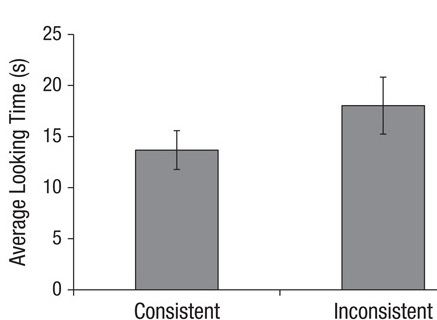}

\caption{
  Infant looking time, a measure of surprise,
  is increased in the inconsistent condition of (c), as compared to the consistent condition, reproduced from \cite{werchan2016role}.
}
\label{fig:cognition4}
\end{subfigure}
\hfill
\begin{subfigure}[b]{.15\linewidth}
\begin{tikzpicture}
\begin{axis}[
    width=100pt,
    height=80pt,
    ybar,
    /pgf/bar width=8pt,
    y label style={align=center,font=\bfseries\scriptsize,yshift=-5pt},
    x label style={font=\bfseries\scriptsize,yshift=5pt},
    tick label style={font=\tiny},
    legend style={
      draw=none,
      at={(0.5,1.15)},
      anchor=south,
      legend columns=-1,
      font=\tiny,
    },
    ylabel={validation loss},
    enlarge x limits=0.4,
    scaled y ticks=false,
    xtick=data,
    symbolic x coords={consistent,inconsistent},
    ]


  \addplot[
    blue,
    error bars/.cd, y dir=both,y explicit,
  ] coordinates {
    (consistent, 3.656) +- (0, 0.19945926902503183)
    (inconsistent, 6.188) +- (0, 0.17127755252805305)
  };


  \addplot[
    red,
    error bars/.cd, y dir=both,y explicit,
  ] coordinates {
    (consistent, 3.908) +- (0, 4.355416857202075)
    (inconsistent, 3.6919999999999993) +- (0, 4.383087496274744)
  };

  \legend{\acrshort{maml-context}, \acrshort{maml}}
\end{axis}
\end{tikzpicture}
\caption{
  Validation set losses ($n$ = 5 random seeds) on a reproduction of the inference test. \acrshort{maml-context} is sensitive to the condition and consistent with human behavior whereas \acrshort{maml} is not.
}
\label{fig:cognition5}
\end{subfigure}
\caption{
  Our proposed method qualitatively reproduces the learning task and inference test from \cite{werchan20158,werchan2016role}, thereby demonstrating the context-sensitive nature of our approach. As a proxy for the looking-time which is a measure of human surprise in \cite{werchan20158,werchan2016role}, we report the training or validation set loss value, which can be interpreted as the degree of surprise of the model towards the data in the training or validation set.
}
\end{figure*}

\section{A meta-learning account of context-specific adaptation in humans}
\label{sec:method}

We now present our framework for capturing context-specific adaptation. In a standard meta-learning setup, a parametric meta-learner encodes information about the shared structure of the distribution of tasks, $p(\task)$, into a set of global parameters $\metaparams$ from which all task-specific predictors are derived. In particular, for each task $\task\taskidx{j} \sample p(\task)$, the meta-learner receives a task-specific dataset $\mathcal{D}_{j}=\left\{\mathbf{x}_{j_{i}}, \mathbf{y}_{j_{i}}\right\}$ and produces a predictive distribution $p_{\metaparams}(\hat{\mathbf{y}}_{j} \given \hat{\mathbf{x}}_{j}, \mathcal{D}\taskidx{j})$ for new examples $\hat{\mathbf{x}}_{j}$ from the same task. 

Here, to capture context-sensitive adaptation, we propose to augment the standard meta-learning problem statement in a way that is analogous to the way contextual cues prime human learning in a new environment. In particular, we posit that the additional environmental contextual information, $\context_j$, can be leveraged as conditioning information in order to prime the initial state of the model $\metaparams$ for a specific task $\task_j$ (also refer to Figure~\ref{fig:architecture}). Formally, we implement conditioning on the task information $\context$ by parameterizing the initialization $\metaparams$ as the output of a context model $g$ with parameters $\contextparams$. Using experience from the task, $\metaparams$ is subsequently adapted with gradient descent to task specific parameters $\taskparams$, as in \gls{maml}. In practice, we take $g$ to be a neural network with weights $\contextparams$, which we refer to as a \emph{context network}, and update $\contextparams$ via back-propagation. Note that $\contextparams$ is updated only during the meta-update step and during the inner loop for task-specific adaptation, $\metaparams$ is used to initialize $\taskparams$ which is subsequently updated based on task-specific data. 

\subsubsection{Supervised meta-learning with task information.} We consider a family of tasks $\task$ with shared structure that enables a meta-learner to learn to solve a task from $\mathcal{T}_i\sim p(\mathcal{T})$. In the supervised learning setting, each task $\mathcal{T}_i$ consists of a set of examples $\mathbf{x}$ and annotations $\mathbf{y}$ (\eg images with classification labels). Gradient-based meta-learning methods choose a parameterized model (base learner) and define an optimization objective over $\mathcal{T}$. For instance, the \gls{maml} algorithm~\cite{finn2017model} uses the following objective:
\begin{equation}
  \min_{\metaparams} \mathbb{E}_{\task_i}\left[\mathcal{L}_{\mathcal{T}_i} (f_{\metaparams'})\right] =  \mathbb{E}_{\task_i}\left[\mathcal{L}_{\mathcal{T}_i} (f_{\metaparams - \alpha \nabla_{\metaparams} \mathcal{L}_{\mathcal{T}_i}(f_{\metaparams}})\right]
\end{equation}
where $f$ is a parameterized function representing the base learner or policy, $\metaparams$ refers to the parameters that are optimized in the outer loop, and the $\taskparams$ parameters are used to compute the objective with respect to each task.
When employing task information, the meta-objective becomes
\begin{equation}\label{eq:maml-ctx}
\min_{\contextparams,\metaparams}\  
\mathbb{E}_{\task_i \sample p(\task)}\left[
\mathcal{L}_{\mathcal{D}^\text{q}\taskidx{i}} \left(
f_{\{ g_{\contextparams}(\context_i),\metaparams - \nabla_{\metaparams}{ [\mathcal{L}_{\mathcal{D}^\text{s}\taskidx{i}}(f_{\{ g_{\contextparams}(\context_i),\metaparams\}})]
} \} }
\right)\right]~,
\end{equation}
where the principal difference is that the initial parameterization of the base network depends not only on global parameters $\metaparams$, but also task-information-dependent parameters produced as the output of $ g_{\contextparams}(\cdot)$. With this meta-objective, we can thus fully differentiate the objective with respect to $\metaparams$; we may make a further application of the chain rule to derive an update for $\contextparams$ also using the objective value at the last inner adaptation step.
 
\subsubsection{Meta-policy search with task information.}
\Glsfirst{rl} assumes a \gls{mdp} consisting of \smash{$(S, A, P, R, \gamma)$}, and the goal is to discover a policy $\policy$ that maximizes the return $\sum_{k=0} \gamma^k R_{k+1}$, the sum of episodic rewards discounted by $\gamma \in [0, 1)$~\citep{sutton2018reinforcement}.
\Glsfirst{meta-rl} generalizes this setting to a distribution $\mdpdistribution$ over \gls{mdp}s, with the aim of finding the policy that maximizes the expectation of returns under this distribution:
$\mathbb{E}_\mdpdistribution\left[\sum_{k=0} \gamma^k R_{k+1}\right]$.
Similar to the supervised scenario, we can decouple a solution to the \gls{meta-rl} problem by performing an outer loop search procedure for parameters that maximized expected return across a distribution of control tasks,
$\mathbb{E}_{\task_i}\left[
\mathcal{L}_{\mathcal{T}_i} (f_{\metaparams})\right] = \mathbb{E}_{\task_i}\left[-\mathbb{E}_{(s_t, a_t) \sim q_{\mathcal{T}_i}} \big[\sum_t R(s_t, a_t)\big]\right]$
where $q_{\mathcal{T}_i}$ is the transition distribution of task $\mathcal{T}_i$, $s_t$ and $a_t$ are state and action at time $t$, respectively. The main difference from the supervised case is that we cannot explicitly differentiate through the dynamics of the environment, and so the standard approach is to use policy gradient methods to update meta-parameters $\metaparams$; we refer to \cite{finn2017model} for more details.  With task information, algorithmically, updating $\psi$ and $\metaparams$ is similar to the supervised case. During the inner adaptation steps, only $\metaparams$ is updated to compute the task-specific parameters $\taskparams$. However, during the meta-update step, the gradient of the post-update objective value is used to update both $\contextparams$ and $\metaparams$, in a generalization of the \gls{maml} algorithm.

\subsubsection{Implementing a context-conditioning network.} 

Learning a function $g$ that produces a parameter initialization for a high-dimensional function $f$ such as a neural network poses problems of under-fitting and computational inefficiency.
There have been methods proposed to alleviate this issue \citep[\eg][]{ha2017hypernetworks,mackay2019self}, all resting on the same premise (or empirical demonstration) that producing a subset of the parameter of $f$ is sufficient.
In all our large-scale experiments, we make use of the \gls{film} parameterization from \cite{perez2018film}; namely, the context network $g$ produces the shift and scale parameters in the hidden layers~\citep{ioffe2015batch} of the base network $f$, thereby acting to linearly transform activations in the base network.

\section{Modeling human behavior}
\label{sec:cognitive}

We begin by applying our proposed framework to capture human behavior in a well-known cognitive control experiment.

\subsubsection{Task description.} We model our setup after the experiment in \cite{werchan20158, werchan2016role}. In their study, 8-month-old infants participated in a learning task followed by a violation-of-expectation inference test. In the learning task, infants viewed toy-word mappings that could be grouped into distinct rule sets using the faces and corresponding voices as higher-order contexts (refer to Figure \ref{fig:cognition1}). Each face-voice context labeled the toys using different words, similar to a bilingual environment in which one caregiver speaks English, and another caregiver, Spanish. Near the end of the learning task, a novel face-voice context was presented with several observed toy-word pairs and a novel toy-word pairing. This is akin to the infant observing a new caregiver introducing a new word in Spanish. During the inference test, infants were presented with the first two face-voice contexts from the learning task paired with the novel toy-word pairing presented at the end of the learning task (refer to Figure ~\ref{fig:cognition3}). One of these presentations was consistent with the rule set structure formed during learning, while the other was inconsistent. Sensitivity to this contrast would demonstrate that the infant infers that the Spanish-speaking caregiver should use the novel object-label mapping introduced by the third caregiver, while the English-speaking caregiver should not. Infants looked longer at the inconsistent pairing compared to the consistent pairing, implying greater surprisal during inconsistent pairings~\footnote{Looking time is a common metric used to study children's cognitive indications such as surprise and expectation violation}. If contextual cues did not help learn a hierarchical rule set, then no difference in the looking time would have been observed. Similar studies have also been undertaken with adults \citep{collins2012reasoning, collins2013cognitive}, demonstrating that both adults and infants use contextual cues for faster task adaptation.

%
%

\begin{figure*}[t]
\centering
\begin{subfigure}{.42\textwidth}
  \centering
  \setlength\tabcolsep{2pt}
  \begin{tabular}{cp{.26\linewidth}cc:cc}
    \multicolumn{3}{c}{\tiny \textbf{support set}} &&& {\tiny \textbf{query item}} \\
    \multicolumn{3}{c}{\resizebox{135pt}{5pt}{\rotatebox[origin=c]{270}{\Bigg[}}} &&&
    \resizebox{30pt}{5pt}{\rotatebox[origin=c]{270}{\Bigg[}} \\
    \includegraphics[width=.15\linewidth]{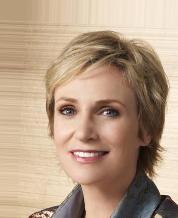} &
    \centering
    \raisebox{12pt}{\parbox[c]{\linewidth}{
      \centering
      \tiny
      {\color{green}\cmark} short hair {\color{red}\xmark} \\
      {\color{green}\cmark} mouth open {\color{red}\xmark} \\
      {\color{green}\cmark} light eyes {\color{red}\xmark}
    }} &
    \includegraphics[width=.15\linewidth]{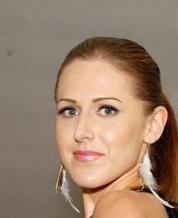} &&&
    \includegraphics[width=.15\linewidth]{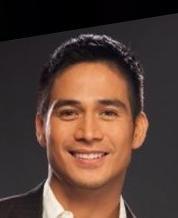} \\
    \parbox[c]{0.2\linewidth}{
      \centering
      \tiny 
      $y=1$
      \\
      (positive ex.)
    } &&
    \parbox[c]{0.2\linewidth}{
      \centering
      \tiny 
      $y=0$
      \\
      (negative ex.)
    } 
    &&&
      \tiny 
      $y=$ ?
  \end{tabular}
  \caption{
    The classification boundary separating the positive and the negative example is ambiguous;
    the query example is correctly classified when only two attributes (\emph{short hair} and \emph{mouth open}) are used for prediction.
  }
  \label{fig:celeba-a}
\end{subfigure}%
\hfill
\begin{subfigure}{.25\textwidth}
\begin{tikzpicture}
\begin{axis}[
    width=120pt,
    height=80pt,
    ybar,
    /pgf/bar width=2pt,
    y label style={align=center,font=\bfseries\scriptsize,yshift=-5pt},
    x label style={font=\bfseries\scriptsize,yshift=5pt},
    tick label style={font=\tiny},
    legend style={
      draw=none,
      at={(0.5,1.15)},
      anchor=south,
      legend columns=-1,
      font=\tiny,
    },
    ylabel={count of examples},
    xlabel={number of attributes},
    xmin=0,
    scaled y ticks=false,
    yticklabels={0,10000,20000,30000},
    ]
  \addplot coordinates {
    (1, 35)
    (2, 483)
    (3, 2424)
    (4, 6103)
    (5, 10613)
    (6, 14833)
    (7, 18057)
    (8, 20106)
    (9, 20198)
    (10, 18983)
    (11, 16488)
    (12, 13368)
    (13, 9556)
    (14, 6103)
    (15, 3331)
    (16, 1409)
    (17, 528)
    (18, 123)
    (19, 24)
    (20, 5)
  };

  \addplot coordinates {
    (1,  5)
    (2,  73)
    (3,  312)
    (4,  768)
    (5,  1370)
    (6,  1848)
    (7,  2255)
    (8,  2642)
    (9,  2414)
    (10, 2288)
    (11, 2085)
    (12, 1543)
    (13, 1149)
    (14, 684)
    (15, 289)
    (16, 110)
    (17, 28)
    (18, 3)
    (19, 1)
    (20, 0)
  };
  \addplot coordinates {
    (1,  5)
    (2,  64)
    (3,  301)
    (4,  712)
    (5,  1169)
    (6,  1674)
    (7,  2069)
    (8,  2272)
    (9,  2488)
    (10, 2277)
    (11, 2181)
    (12, 1793)
    (13, 1238)
    (14, 863)
    (15, 495)
    (16, 239)
    (17, 79)
    (18, 34)
    (19, 8)
    (20, 1)
  };
  \legend{\texttt{train},\texttt{valid},\texttt{test}}
\end{axis}
\end{tikzpicture}
  \caption{
    The median \acrshort{celeba} image has 8 to 9 attributes active, and so 
    the task in (a) is often ambiguous.
}
  \label{fig:celeba-b}
\end{subfigure}%
\hfill
\begin{subfigure}{.26\textwidth}
\vfill
\centering
\setlength{\tabcolsep}{2pt}
\scriptsize
\begin{tabular}{l|cc}
\toprule
 & Meta-test acc. \\
\midrule
  \acrshort{maml}    & 68.61\% $\pm$ 3.71\% \\
  \gls{maml-static}  & 63.56\% $\pm$ 4.00\% \\
  \gls{maml-concat}  & 75.61\% $\pm$ 3.46\% \\
  \acrshort{maml-context} & \textbf{77.51\% $\pm$ 3.34\%} \\
\bottomrule
\end{tabular}
\caption{
  Average accuracy and 95\% confidence interval over 600 evaluation episodes from the 
  meta-testing split (held-out images and attribute combinations) of \acrshort{celeba}.
}
  \label{tab:celeba-c}
\end{subfigure}
\caption{
  The \glsfirst{celeba} ambiguous classification task set:
  When more than $k$ attributes are consistent with a classification of the support set,
  the $k$-shot classification task is ambiguous (see (a));
  the \glsfirst{celeba} task set is such a setting (see (b))
  and so requires task information in order to prime the classification decision on a query item,
  as demonstrated in (c).
}
\label{fig:celeba}
\end{figure*}
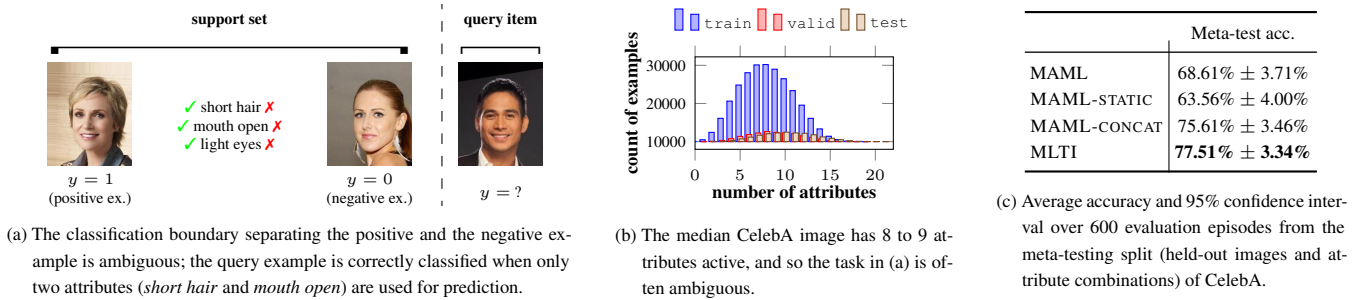

\subsubsection{Experimental setup.} If our framework can capture context-sensitive adaptation, then we should be able to replicate the looking-time results from \cite{werchan20158, werchan2016role}. To test this, we created an analogous problem setting which consisted of a similar leaning task and inference test. During the learning task, we provided  tasks comprising a context, $\context \in \{0, 1, 2\}$ representing the speaker identity and two disjoint batches of stimulus-response pairs $(\mathbf{x}, \mathbf{y}) \in \{0, 1, 2\} \times \{0, 1, 2, 3, 4\}$, each representing an object identity paired with a word label. Like in the behavioral learning task, stimulus-response mappings appear only within valid contexts. Further, one of the stimulus-response pairs,
$(\mathbf{x}, \mathbf{y}) = (2, 4)$ is only presented in one context ($\context=2$) even though it is valid in another ($\context=0$). For the inference test, we create two conditions -- consistent and inconsistent. In the consistent condition, the context network is presented with context $\context=0$, the produced parameters are adapted with seen examples from the context, and the adapted model's loss is evaluated on the held-out stimulus-response mapping, $(\mathbf{x}, \mathbf{y}) = (2, 4)$. In the inconsistent condition, the context network is presented with context $\context=1$, the produced parameters are adapted with seen examples from the context, and the adapted model's loss is evaluated on the held-out stimulus-response mapping, $(\mathbf{x}, \mathbf{y}) = (2, 4)$. Detailed data sampling procedure and worked-out task examples are included in the Supplementary. 

\subsubsection{Hyperparameters.} Both the base and context networks use a neural network with two hidden layers of size 10. Since $\metaparams$'s dimensionality is sufficiently low, the context network, which maps task information to base network weights, directly outputs $\metaparams$. For task-specific adaptation, we use one gradient update using a support set of size $10$. During inference, we present the model with $2$ support examples from a newly sampled task and measure mean-squared error over $1$ query example.

\subsubsection{Results.} We compare our approach, which we term \gls{maml-context}, against \gls{maml}, the meta-learning method for \gls{sl} in \citep{finn2017model}. We hypothesize that our framework should be sensitive to the evaluation condition just like humans. Since \gls{maml-context} uses the context as higher-order information, its error in the consistent condition should be lower compared to the error in the inconsistent condition (analogous to the difference in looking time/surprise in humans). On the other hand, because \gls{maml} doesn't utilize contextual information, its error should not be influenced by the condition. Thus, its performance would serve as an ideal baseline to compare our framework. Note that absolute value of the validation errors are not particularly important, rather the relative difference in the validation errors across conditions is more important. 

We first see that \gls{maml-context} learns faster compared to \gls{maml} (Figure~\ref{fig:cognition2}). This is not surprising as \gls{maml-context} employs the contextual information whereas \gls{maml} does not. We further note that the variance in the performance of \gls{maml} is quite high. Next, in Figure~\ref{fig:cognition5} we see that \gls{maml-context} can qualitatively reproduce the looking time results from \citep{werchan20158, werchan2016role} as the error of \gls{maml-context} in the consistent condition is considerably lower than the error in the inconsistent condition ($3.65$ vs. $6.19$). A paired t-test revealed that this difference was statistically significant, $t(8) = -19.3, p < 0.001$. We further observe that as per our predictions, the baseline \gls{maml} is not affected by the difference in condition as its error on the consistent condition is similar to the inconsistent condition ($3.9$ vs $3.7$). A paired t-test revealed that this difference was not significant, $t(8)  = 0.1, p = 0.5$. We also observe that the variance in the error of \gls{maml} is quite high. This is partly driven by the high variance during learning -- whenever \gls{maml} reaches a lower error on the meta-training, it overfits on the training set (due to lack of context information) leading to a very high loss on the validation set. These results show that predictions made by our proposed framework are consistent with human behavior in a well-studied cognitive control task. 

\section{Large-scale experiments}
\label{sec:expoverview}

The previous section showed that \glsfirst{maml-context} is consistent with psychological findings about context-sensitive adaptation on a controlled cognitive task. We now evaluate whether \gls{maml-context} can perform competitively in more complex problem settings by guiding adaptation in meta-learning.

\subsubsection{Overview of task information.} In the setting of \acrshort{mujoco}, we explore task information as a diagnostic cue by using scalar parameter as task information. For the more challenging \acrshort{celeba} dataset, we use a binary vector with attribute information as task information. For the \gls{metaworld} tasks, we use the 3D goal position as task information.

\subsubsection{Baseline comparisons.} We compare our approach of context-conditioned adaptation, \gls{maml-context}, against three categories of baseline as described below. For hyperparameters that are common to all comparison methods, we use the same settings as are used in \cite{finn2017model} and \cite{rothfuss2019promp} where applicable. 

\textbf{\Gls{maml}} is the meta-learning method for \gls{sl} as described in \citep{finn2017model} and \textbf{\gls{promp}} is a policy-gradient meta-\gls{rl} method that improves upon the initial application of \gls{maml} to \gls{rl} in \cite{finn2017model} by combining \gls{maml} and \gls{ppo} \cite{rothfuss2019promp}.
These methods make no use of task information and serve as lower bounds to task-information conditioning.

\textbf{\Gls{maml-static}} and \textbf{\gls{promp-static}} are baselines with the same architecture as \gls{maml-context} but do not depend on the context and instead replace the context $\context$ with a constant vector; this baseline is intended as a parameter count-equivalent baseline to \gls{maml-context} in order to distinguish architectural differences in performance as compared to \gls{maml} and \gls{promp}.
 
\textbf{\Gls{maml-concat}} and \textbf{\gls{promp-concat}} use the same architecture as the \gls{maml-context} method but use task information in the form of concatenation to the observation; this setup is analogous to goal-conditioned RL, where policies are trained to reach a goal state that is provided as additional input~\cite{kaelbling1993learning,schaul2015universal,pong2018temporal,sutton2019horde}.
These baselines are provided with the same amount of information as \gls{maml-context} but do not decouple context and task-specific feedback into initialization and adaptation phases, respectively, and therefore test the utility of task-information in priming meta-learning like humans do as opposed to simply being treated as extra observational information.


\begin{figure*}[t]
\setlength{\tabcolsep}{3pt}
  \centering
\begin{tabular}{cccc}

\raisebox{12pt}{
\includegraphics[height=38pt]{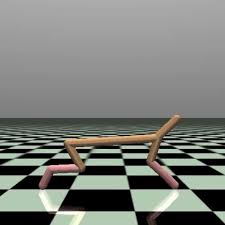}
}

&

\begin{tikzpicture}%
    \begin{axis}[%
      avg-return,
      xmin=0, xmax=99999999,
      ylabel={average return},
      ymin=-5, ymax=350,
      ytick={0,50,100,150,200,250,300,350},
      scaled x ticks=true,
      every x tick label/.append style={alias=XTick,inner xsep=0pt},
      every x tick scale label/.style={at=(XTick.base east),anchor=base west}
    ]

    \errorband{data/mujoco/halfcheetah-none.csv}{iteration}{mean}{std}{black}{0.1}
    \errorband{data/mujoco/halfcheetah-static.csv}{iteration}{mean}{std}{black!15!green}{0.1}
    \errorband{data/mujoco/halfcheetah-concat.csv}{iteration}{mean}{std}{blue}{0.1}
    \errorband{data/mujoco/halfcheetah-dynamic.csv}{iteration}{mean}{std}{orange}{0.1}

    \end{axis}
\end{tikzpicture}%

&

\begin{tikzpicture}%
    \begin{axis}[%
      avg-return,
      xmin=0, xmax=99999999,
      ymin=100, ymax=350,
      ytick={100,150,200,250,300,350},
      scaled x ticks=true,
      every x tick label/.append style={alias=XTick,inner xsep=0pt},
      every x tick scale label/.style={at=(XTick.base east),anchor=base west}
    ]

    \errorband{data/mujoco/walker-none.csv}{iteration}{mean}{std}{black}{0.1}
    \errorband{data/mujoco/walker-static.csv}{iteration}{mean}{std}{black!15!green}{0.1}
    \errorband{data/mujoco/walker-concat.csv}{iteration}{mean}{std}{blue}{0.1}
    \errorband{data/mujoco/walker-dynamic.csv}{iteration}{mean}{std}{orange}{0.1}

    \end{axis}
\end{tikzpicture}%

&

\begin{tikzpicture}%
    \begin{axis}[%
      avg-return,
      xmin=0, xmax=99999999,
      ymin=-300, ymax=-150,
      ytick={-300,-250,-200,-150},
      scaled x ticks=true,
      every x tick label/.append style={alias=XTick,inner xsep=0pt},
      every x tick scale label/.style={at=(XTick.base east),anchor=base west},
      legend style={at={(1.4,.5)},anchor=west},
    ]

    \errorband{data/mujoco/ant-none.csv}{iteration}{mean}{std}{black}{0.1}
    \addlegendentry{\acrshort{promp} (base)}
    \errorband{data/mujoco/ant-static.csv}{iteration}{mean}{std}{black!15!green}{0.1}
    \addlegendentry{\gls{promp-static} \\ (ablation)}
    \errorband{data/mujoco/ant-concat.csv}{iteration}{mean}{std}{blue}{0.1}
    \addlegendentry{\gls{promp-concat} \\ (strong baseline)}
    \errorband{data/mujoco/ant-dynamic.csv}{iteration}{mean}{std}{orange}{0.1}
    \addlegendentry{\acrshort{promp-context} (\textbf{ours})}

    \end{axis}
\end{tikzpicture}%

\\
&
\multicolumn{3}{c}{
  \hspace{-70pt}
  \scriptsize \textbf{number of timesteps} (millions)}

\end{tabular}
\caption{
    From left to right, performance of the different methods on the 
    \gls{half-cheetah},
    \gls{walker-2d}, and
    \gls{ant-goal} environments
    from \acrshort{mujoco}.
    For each method, we report the mean and standard deviation over three random seeds.
}
\label{fig:mujoco}
\end{figure*}

\subsection{Ambiguous classification with \acrshort{celeba}}
\label{sec:celeba}

\subsubsection{Experimental setup.}
We use a construction similar to \cite{finn2018probabilistic} to generate an ambiguous binary classification task with the \gls{celeba} dataset.
\footnote{We focused on the \gls{celeba} dataset instead of \gls{mini}, the usual dataset for the evaluation of few-shot classification methods~\cite{vinyals2016matching}, as we can easily generate task descriptors.}
In particular, for each task, we sample 2 of the 40 attributes from \gls{celeba}, then subsequently sample for the support set one image that contains these attributes (a positive example) and one image that does not contain these attributes (a negative example); this binary classification task is often ambiguous, as most images in \gls{celeba} have more than two attributes active.
The task information is provided in the form of a two-hot vector that identifies the two attributes upon which the base network has to make a classification decision.
The query set comprises 15 examples as in the experimental setup in \cite{vinyals2016matching}.

\subsubsection{Hyperparameters.} 
The context network pipeline embeds the two-hot task information vector via a learned embedding matrix; these embeddings are summed then fed as input to a two-layer feed-forward neural network with 40 hidden units.
As per the implementation of \gls{film}-conditioning, the context network outputs a feature map that performs linear transformations to the base network's hidden activations.
The base network itself is a four-layer convolutional network with 32 filters applied at stride 2, similar to the small-scale convolution network employed in few-shot classification on the \gls{mini} dataset~\citep{vinyals2016matching,finn2017model}.
We set hyperparameters on the held-out validation set; all settings as well as details on the implementation of the context network are included in Supplementary.

\subsubsection{Results.} As shown in Table 1, \gls{maml-static} suffers from the need to fit the extra parameters and \gls{maml} performs the task with a low degree of accuracy.
Next, we see that \gls{maml-context} performs marginally better than \gls{maml-concat}. These results suggests that for the highly-ambiguous few-shot \gls{celeba} task, our cognitively-inspired method outperforms the context-independent method like \gls{maml} while performing competitively (if not better) compared to context-concatenation method \gls{maml-concat}.

\bgroup
\setlength{\tabcolsep}{0pt}
\begin{figure*}[t]

\begin{tabular}{c}

\foreach \i/\env/\ylabel/\colwidth/\hspacebefore in {
    1/reach/average return/90pt/16pt,
    2/door-lock/ /85pt/10pt,
    3/door-unlock/ /85pt/10pt,
    4/soccer/ /85pt/10pt,
    5/basket-ball/ /85pt/-8pt}
  {
    \begin{tabular}{p{\colwidth}}
      \centering
      \hspace{\hspacebefore}
      \includegraphics[width=40pt]{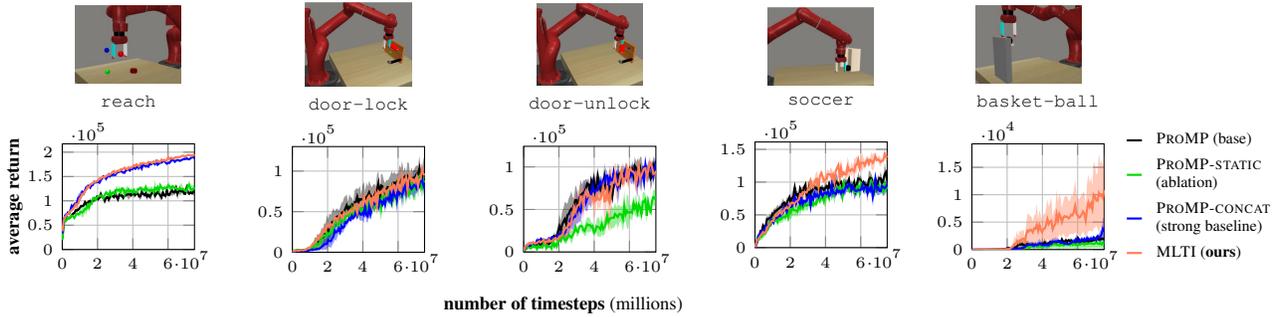} \\
      \begin{tikzpicture}%
        \begin{axis}[%
          avg-return,
          xmin=0, xmax=75000000,
          ymin=0, 
          ylabel={\ylabel},
          title=\texttt{\scriptsize\env},
          title style={yshift=1ex,},
          scaled x ticks=true,
          every x tick label/.append style={alias=XTick,inner xsep=0pt},
          every x tick scale label/.style={at=(XTick.base east),anchor=base west}
        ]

        \errorband{data/metaworld/\env-none.csv}{iteration}{mean}{std}{black}{0.4}
        \ifnum \i=5
          \addlegendentry{\acrshort{promp} (base)}
        \fi

        \errorband{data/metaworld/\env-static.csv}{iteration}{mean}{std}{black!15!green}{0.4}
        \ifnum \i=5
          \addlegendentry{\gls{promp-static} \\ (ablation)}
        \fi

        \errorband{data/metaworld/\env-concat.csv}{iteration}{mean}{std}{blue}{0.4}
        \ifnum \i=5
          \addlegendentry{\gls{promp-concat} \\ (strong baseline)}
        \fi

        \errorband{data/metaworld/\env-dynamic.csv}{iteration}{mean}{std}{orange}{0.4}
        \ifnum \i=5
          \addlegendentry{\acrshort{promp-context} (\textbf{ours})}
        \fi

        \end{axis}
    \end{tikzpicture}
  \end{tabular}}

\\

{\scriptsize \textbf{number of timesteps} (millions)}

\end{tabular}
\caption{
    From left to right, 
    performance, measured by averaging post-update performance over a batch of tasks, on the
    \gls{reach},
    \gls{door-lock},
    \gls{door-unlock},
    \gls{soccer}, and
    \gls{basket-ball}
    environments
    from \gls{metaworld}~\cite{yu2019meta}. 
    For each method, we report the mean and standard deviation over three random seeds.
}
\label{fig:metaworld}
\end{figure*}
\egroup

\subsection{Parameterized \acrshort{mujoco} Tasks}
\label{sec:mujoco}

We next comparing the above methods on simple continuous control tasks by using a set of three parameterized environments from the \gls{mujoco} simulator~\citep{todorov2012mujoco}.
For the below results, the average return for the pre- and post-task-specific adaptation is computed from trajectories sampled before and after the inner loop for task-specific adaptation.
 
\subsubsection{Environments.} 
In these environments, the underlying dynamics of the \gls{half-cheetah}, \gls{walker-2d}, and \gls{ant-goal} environments depend on a randomly sampled scalar parameter: In \gls{half-cheetah} and \gls{walker-2d}, a scalar parameter controls the direction of motion (forward or backward) that is produced for a given action; for \gls{ant-goal}, a randomly sampled 2D position defines a goal to which the actuator must be moved.
We use this scalar parameter as task information for this setting.

\subsubsection{Hyperparameters.} 
For all method including the \gls{promp} and \gls{promp-concat} baselines, the base policy is a fully-connected network with two hidden layers of dimension $64$ and ReLU nonlinearities as in \cite{rothfuss2019promp}.
For \gls{promp-context} and the \gls{promp-static} baseline, the base policy is conditioned with a \gls{film} module; this module is fed contextual input and outputs a feature map that performs linear transformations on the policy network's hidden layers.
In our experiments, \gls{film} is represented as a fully connected network with two hidden layers (of increasing dimension--32 and 64--to achieve up-sampling of the context) and outputs $W_i$ and $b_i$ for each hidden representation $h_i$ in the policy network, performing the transformation $h'_i = W_i \odot h_i + b_i$.
For other hyperparameters that are common to all four comparison methods, we use the same settings as are used in \cite{rothfuss2019promp}.
In particular, the number of inner optimization steps is set to one, entailing two rollouts in the environment to evaluate the pre- and post-adaptation objectives.
 
\subsubsection{Results.} 
Figure \ref{fig:mujoco} reports the post-adaptation performance of all methods.
First, task information is beneficial as \gls{promp-context} consistently outperforms both \gls{promp} and \gls{promp-static} in all three environments.
Next, \gls{promp-context} performs better than \gls{promp-concat} in the \gls{half-cheetah} environment and at least as well as \gls{promp-concat} in \gls{walker-2d} and \gls{ant-goal}.
This suggests that our cognitively inspired approach of learning conditionally with the task information is a competitive parameterization compared to learning jointly (\ie \gls{promp-concat}, a standard in goal-conditioned \gls{rl} setups).

\subsection{\gls{metaworld} manipulation tasks}
\label{sec:metaworld}

We next investigate on a more challenging setting, using a set of five parameterized environments from the \gls{metaworld} benchmark~\citep[][ Figure~\ref{fig:metaworld} (top)]{yu2019meta}.

\subsubsection{Environments.} 
The \gls{metaworld} benchmark presents a variety of simulated continuous control tasks that consist of a 4-dimensional robot actuator that is required to move from an initial position to a goal position.
In these environments, we use the goal position, a 3$\times$1 vector, as  task information; this goal position is re-sampled when a new task is encountered.
Similar to the \gls{mujoco} environments, the goal position is normally treated as a direct concatenation to the state observation; we instead use the goal position as an input to context module in order to investigate the effect of context-conditional adaptation in a meta-policy search algorithm.
This use has two advantages: First, the goal position is readily available in different environments in \gls{metaworld}.
Second, and more importantly, we hypothesize that goal information serves as an integral cue for reducing task ambiguity even in the various dense-reward environments in \gls{metaworld}.
Furthermore, a task is defined as a fixed goal with varying initial states.
This makes these environments more challenging: rather than fixing the goal and initial state, the pre- and post-adaptation policies are evaluated with different initial states and goals.

\subsubsection{Hyperparameters.} 
We use the same base policy and context network implementations as in the previous section.
Since \gls{metaworld} environments are substantially more difficult to solve as compared to the \gls{mujoco} environments, these environments needed more inner adaptation steps to show post-adaptation improvement; an inventory of hyperparameter settings are provided in Supplementary.

\subsubsection{Results.} 
Results are shown in Figure~\ref{fig:metaworld}.
We first observe that even for \gls{reach}, a very simple environment, task information is necessary to perform well on the task, as evident by the superior performance of \gls{promp-context} and \gls{promp-concat}.
One possible explanation for this is that the reward available in \gls{reach} is insufficient to guide meta-learning by itself, and that the goal information serves as a useful cue to guide meta-learning.
Additionally, we observe that in the \gls{reach} environment,  context-conditioning is not especially beneficial compared to context-concatenation as both \gls{promp-context} and \gls{promp-concat} perform similarly on this task.
 
Next, we see that in both \gls{door-lock} and \gls{door-unlock}, task information is not necessarily crucial to perform well, as both \gls{promp-context} and \gls{promp-concat} perform similarly to \gls{promp}.
Interestingly, the over-parameterized architecture in the \gls{promp-static} method worsens the performance on the \gls{door-unlock} environment.
The most interesting cases are the \gls{soccer} and \gls{basket-ball} environments: Here, we see that \Gls{promp-context} significantly outperforms all other methods.
Furthermore, we see that simply providing task-information as an extra input is not beneficial as evident from the performance of the \Gls{promp-concat} method on these two environments.
 
In summary, our proposed contextual-conditional meta-learning outperforms all the methods (including \gls{promp-concat}) on both the \gls{soccer} and \gls{basket-ball} environments and performs as well as the other methods (if not better) in the remaining environments.
As a main takeaway, results from this experiment suggest that our cognitively inspired framework is a promising way to improve the performance of meta-learning on more challenging tasks such as \gls{metaworld}.
 



\section{Conclusion}
\label{sec:conclusion}

An extensive literature in psychology and neuroscience demonstrates that context-specific adaptation is an integral component of cognitive control~\citep{monsell1996control, dosenbach2006core, sakai2008task}. Here, we explain context-sensitive adaptation under a meta-learning framework that integrates task information to guide adaptation to new tasks. Our modeling results on a cognitive control task support existing theories that propose higher-order contextual information helps humans structure learning~\citep{collins2012reasoning, frank2012mechanisms, collins2013cognitive, donoso2014foundations, eckstein2020computational}. According to these theories, hierarchical learning based on contextual cues ensures that learning new information does not conflict with behaviors learned in other contexts; for instance, an infant in a bilingual environment receiving two different labels for the same word would not get confused when labels are consistent with the higher-order context provided by the identity of the speaker.
Our large-scale experiments further show that our cognitively inspired meta-learning framework is also a promising approach towards improved adaptation in meta-learning. Analogous to the way people use contextual cues as a prior over task structure, our framework thus highlights the value of task information in bringing meta-learning algorithms closer to human-like learning.

\section{Ethics Statement}
Our research contributes towards improving our understanding of cognitive control in humans as well as the field of meta-learning, which aims to emulate the ability of humans to learn new tasks rapidly. There are many benefits to such contributions, such as the development of automated systems that can quickly adapt and learn to solve a variety of tasks, although the current problem settings are simplistic as compared to the everyday variability that humans face. However, in the longer term, progress in adaptable and robust algorithms leads towards automation, which will disrupt the labor structures that many people rely on for employment.

\begin{small}
\bibliography{bibs/self,bibs/general,bibs/games,bibs/rl,bibs/bayesian,bibs/communication,bibs/clustering,bibs/cognitive_science,bibs/few_shot,bibs/hier_class,bibs/implementation,bibs/language,bibs/meta_learning,bibs/nlp,bibs/one_class,bibs/optimization,bibs/stats,bibs/vision,bibs/transfer,bibs/karthik}

\end{small}

\appendix
\section{Extended related work}
\label{sec:litrev}

\paragraph{Context-conditioning in meta-learning.}

We now briefly review prior work on context-conditioned meta-learning before contrasting our contribution against these papers.

\acrshort{tadam}~\citep{oreshkin2018tadam} uses \gls{film} conditioning to improve the performance of feature extractors for few-shot classification. However, \Acrshort{tadam} uses a very specific variable for conditioning -- the mean of the class prototypes is the output from a conditional network which serves as inferred task context. On the other hand, we employ \gls{film} with a variety of task-specific contextual cues and show how this can capture context-sensitive adaptation in humans. From our perspective, \Acrshort{tadam} serves as a useful starting point towards considering how context-conditioned meta-learning can be adapted to capture human-like behavior. 

\citep{lee2019learning} learn task-specific balancing variables in order to balance between the meta-knowledge and task-specific update. However, in contrast to our work, they do not employ contextual cues. 

\citep{baik2019learning} aim to control the influence of prior knowledge for each task and propose a method that performs selective forgetting by applying a task-dependent layer-wise attenuation on MAML initialization. This is in contrast to our proposal of utilizing the additional information provided by contextual cues to capture human behavior. 

\citep{vuorio2018toward} is similar to our work from an architectural point of view as it employs a modulation network that produces a task embedding which is used to generate parameters that modulate the task network. However, the key difference is that while \citep{vuorio2018toward} generates the parameters by identifying the mode of tasks sampled from a multi-modal task distribution, we generate the parameters by utilizing contextual information. Future work could investigate the benefits of generating the parameters by utilizing both the multi-modal distribution as well as the auxiliary contextual information. 

\cite{chen2019adaptive} show that using image tags as auxiliary information helps to learn a better representation for prototypical networks, enabling better generalization to unseen tasks. \cite{pahde2018cross} learn both an image classifier and a text-conditioned image generator as a pre-training step; the generator is then used to provide auxiliary data during the few-shot adaptation stage. Both these approaches use contextual information at the instance or class level; in contrast, we operate over task-specific contexts, thus enabling to model human behavior.

\cite{lee2018gradient} is based on the idea that task-specific learning should require fewer degrees of freedom compared to meta-learning and proposes a meta-learning model that determines a subspace and a corresponding metric that task-specific learners can learn in. This is in contrast with our main idea of contextual adaptation. 

\cite{yoon2019tapnet} linearly transform the network output with a task-specific projection; whereas we use contextual information to initialize the meta-learner.

\cite{rakelly2019efficient} learn a policy that adapts to the task at hand by performing inference over a latent context variable on which the policy is conditioned. Here, context is defined as the history of past transitions, which is orthogonal to our setting of using the extra available contextual cues (and not the history of past transitions) to prime learning. Further, they do not investigate priming learning with context variables. 

Lastly, the work of \cite{andreas2018learning} uses auxiliary contextual information to constrain adaptation which makes it closer to our proposed method. However, while \cite{andreas2018learning} perform task-specific parameter estimation in a linguistically structured latent space, we condition on arbitrary task information before interaction with a task, therefore combining more flexible adaptation of task-specific models with guidance provided by arbitrary (\ie beyond linguistic) context. 

\paragraph{Other uses of context.} In addition to context-conditioned meta-learning, there has been a wide variety of work that study the utility of contextual information in decision-making. In the supervised setting, the use of descriptions or tags as extra inputs improves fine-grained image classification~\citep{reed2016learning,he2017fine} and zero-shot learning~\citep{norouzi2013zero}. Contextual information has also been used in sequential decision-making in the form of instruction following~\citep{macmahon2006walk,vogel2010learning,branavan2010reading,chen2011learning,artzi2013weakly,kim2013adapting,Andreas15Instructions}, to guide learning of reward functions~\citep{bahdanau2018learning,zou2019reward} and environment models~\citep{narasimhan2018grounding}, or for better exploration~\citep{harrison2018guiding}. While these methods make use of contextual information, they do so in parallel with concept or policy learning and usually do not deal with few-shot settings. This is analogous to the \textsc{concat} baseline used in our experiments and therefore cannot capture context-specific adaptation in humans. Here, we use contextual information to guide the initialization of task-specific parameters, followed by few-shot adaptation using feedback from the target task; this ordering enforces the use of the task information as a prime for interaction with the target task, similarly to context-sensitive adaptation in humans.

\clearpage

\section{Additional experimental details}
\label{sec:exp_details}

\subsection{Modeling human behavior}

For the cognitive modeling experiment, we report the average of five seeds. 

During the learning task, to reproduce the behavioral task of \cite{werchan20158, werchan2016role}, we provided  tasks comprising a context, $\context \in \{0, 1, 2\}$ and two disjoint batches of stimulus-response pairs $(\mathbf{x}, \mathbf{y}) \in \{0, 1, 2\} \times \{0, 1, 2, 3, 4\}$, where each stimulus-response mappings appeared only within valid contexts. Table \ref{tab:cog} presents the training data sampling procedure in detail.

The hyperparameters are provided below. Further details can be determined by inspecting the attached code that reproduces all of our results (\texttt{code\_cognitive.zip}).

\begin{tabular}{l|ccc}
\toprule
{\sc Cognitive} & {\sc Hyperparameters}  \\
\midrule
Gradient Clip Norm & 10.0 \\
Inner Loop Learning Rate & $0.1$ \\
Outer Loop Learning Rate & $0.005$ \\
Number of Meta-training Steps & $100$\\
Number of Inner Adaptation Steps & 1 \\
\bottomrule
\end{tabular}

\subsection{\gls{celeba}}

The hyperparameters for the \gls{celeba} experiments are provided below.
Note that \cite{finn2018probabilistic} hold out entire attributes at meta-test time, while we hold out combinations of attributes; our setup therefore treats the \gls{celeba} attributes similarly to natural language descriptions with no unobserved vocabulary. An interesting next step would be to add in a component that extrapolates the context network to be applied to out-of-vocabulary items~\citep[\eg][]{hice2019}.
Further details can be determined by inspecting the attached code that reproduces all of our results (\texttt{code\_celeba.zip}).

\begin{tabular}{l|ccc}
\toprule
{\sc \gls{celeba}} & {\sc Hyperparameters}  \\
\midrule
Gradient Clip Norm & 10.0 \\
Inner Loop Learning Rate & $0.01$ \\
Outer Loop Learning Rate & $0.001$ \\
Number of Meta-training Steps & $10^{4}$\\
Number of Inner Adaptation Steps & 5 \\
Meta-batch Size & 4 \\
\bottomrule
\end{tabular}

\subsection{Reinforcement learning experiments}

For all \gls{rl} experiments, we report the average over three seeds. The hyperparameters for \gls{mujoco} and \gls{metaworld} are provided below.
Further details (and the environment-specific horizon length) can be determined by inspecting the attached that reproduces all of our results (\texttt{code\_rl.zip}).

\begin{tabular}{l|ccc}
\toprule
{\sc \gls{mujoco}} & {\sc Hyperparameters}  \\
\midrule
Clip Parameter & 0.3 \\
Discount ($\gamma$) & 0.99 \\
Lambda ($\lambda$)  & 1.0 \\
KL Coeff & 0.0 \\
Learning Rate & $3.0\cdot 10^{-4}$ \\
Tasks per Iteration & 40 \\
Trajectories per Task & 20 \\
Inner Step Size $\alpha$ & 0.1 \\
Inner Adaptation Steps & 1-2 (env-specific) \\
Grad Steps Per \gls{promp} Iter & 3-5 (env-specific) \\
\bottomrule
\end{tabular}

\begin{tabular}{l|ccc}
\toprule
{\sc \gls{metaworld}} & {\sc Hyperparameters}  \\
\midrule
Clip Parameter & 1.0 \\
Discount ($\gamma$) & 0.99 \\
Lambda ($\lambda$)  & 1.0 \\
KL Coeff & 0.0 \\
Learning Rate & $3.0\cdot 10^{-4}$ \\
Tasks per Iteration & 20 \\
Trajectories per Task & 5 \\
Inner Step Size $\alpha$ & 0.05 \\
Inner Adaptation Steps & 4 \\
Grad Steps Per \gls{promp} Iter & 5 \\
\bottomrule
\end{tabular}

\section{Architecture details}

Table \ref{tab:architecture} provides the architecture details for the different experiments. Note that FC(x,y) is a standard fully-connected network with two hidden layers of size x and y, Conv([x,y], s, f, n) is a n layer convolutional network with f kernels of  size [x,y] with stride length s, and LSTM([x, y], h) is a LSTM network with hidden layers of size x and y with a hidden state of size h.

\begin{table*}[htb]
\centering
\begin{tabular}{l|ccc}
\toprule
{\sc Meta-train} & {\sc Meta-test}  \\
\midrule
10 points with $\context = 0, \mathbf{x} = 0, \mathbf{y} = 0$ & 2 points with $\context = 0, \mathbf{x} = 1, \mathbf{y} = 1$ \\
10 points with $\context = 0, \mathbf{x} = 1, \mathbf{y} = 1$ & 2 points with $\context = 0, \mathbf{x} = 0, \mathbf{y} = 0$ \\
10 points with $\context = 1, \mathbf{x} = 0, \mathbf{y} = 2$ & 2 points with $\context = 1, \mathbf{x} = 1, \mathbf{y} = 3$ \\
10 points with $\context = 1, \mathbf{x} = 1, \mathbf{y} = 3$ & 2 points with $\context = 1, \mathbf{x} = 0, \mathbf{y} = 2$ \\
5 points with $\context = 2, \mathbf{x} = 0, \mathbf{y} = 0$; 5 with $\context = 2, \mathbf{x} = 1, \mathbf{y} = 1$; & 2 points with $\context = 2, \mathbf{x} = 2, \mathbf{y} = 4$ \\
5 points with $\context = 2, \mathbf{x} = 0, \mathbf{y} = 0$; 5 with $\context = 2, \mathbf{x} = 2, \mathbf{y} = 4$; & 2 points with $\context = 2, \mathbf{x} = 1, \mathbf{y} = 1$ \\
5 points with $\context = 2, \mathbf{x} = 1, \mathbf{y} = 1$; 5 with $\context = 2, \mathbf{x} = 2, \mathbf{y} = 4$; & 2 points with $\context = 2, \mathbf{x} = 0, \mathbf{y} = 0$ \\
\bottomrule
\end{tabular}
\caption{Detailed training procedure for the cognitive modeling experiment}
\label{tab:cog}
\end{table*}

\begin{table*}
\centering
\begin{tabular}{l|p{2.5cm}p{2.5cm}p{2.5cm}p{2.7cm}p{2.5cm}}
\toprule
{\sc Dataset} & {Base Network} & {Context Network} & {\sc*-static Input} & {\sc*-concat Input} & {\sc \gls{maml-context} Input}  \\
\midrule
\gls{celeba} & Conv([3,3], 2, 32, 4) & FC(40, 40) with \gls{film} conditioning & Constant Vector embedded by a LSTM([40,40], 32) & Two-hot vector w/ attribute information & Two-hot vector w/ attribute information embedded by a LSTM([40,40], 40) \\
\hline
\\
\gls{mujoco},\gls{metaworld} & FC(64, 64) & FC(32, 64) & Constant Vector & Scalar parameter for \gls{mujoco}, 3D goal position for \gls{metaworld}) & Scalar parameter for \gls{mujoco}, 3D goal position for \gls{metaworld})\\
\bottomrule
\end{tabular}
\caption{Architectural details for the experiments. The first two columns correspond to the network architecture for the base and contextual network respectively. The last three columns describe the type of contextual input that is fed into the context network for Static, Concat, and MLTI baselines. Note that for \gls{maml} and \gls{promp} baselines, there is no contextual input. }
\label{tab:architecture}
\end{table*}



\end{document}